\documentclass[a4paper,conference]{IEEEtran}

\usepackage{graphicx}
\usepackage{cite}

\ifCLASSINFOpdf
\else
\fi
\usepackage{amsmath}
\usepackage{amssymb}
\usepackage{algorithmic}
\usepackage{algorithm}
\begin{document}

\title{Spiking Neural Networks with Single-Spike Temporal-Coded Neurons for Network Intrusion Detection}

\author{\IEEEauthorblockN{Shibo Zhou and Xiaohua Li}
\IEEEauthorblockA{Department of Electrical and Computer Engineering\\
State University of New York at Binghamton\\
Binghamton, NY 13902 \\
Email: \{szhou19, xli\}@binghamton.edu}}
%\and
%\IEEEauthorblockN{James Kirk\\ and Montgomery Scott}
%\IEEEauthorblockA{Starfleet Academy\\
%San Francisco, California 96678--2391\\
%Telephone: (800) 555--1212\\
%Fax: (888) 555--1212}}

%
%\author{\IEEEauthorblockN{Michael Shell\IEEEauthorrefmark{1},
%Homer Simpson\IEEEauthorrefmark{2},
%James Kirk\IEEEauthorrefmark{3},
%Montgomery Scott\IEEEauthorrefmark{3} and
%Eldon Tyrell\IEEEauthorrefmark{4}}
%\IEEEauthorblockA{\IEEEauthorrefmark{1}School of Electrical and Computer Engineering\\
%Georgia Institute of Technology,
%Atlanta, Georgia 30332--0250\\ Email: see http://www.michaelshell.org/contact.html}
%\IEEEauthorblockA{\IEEEauthorrefmark{2}Twentieth Century Fox, Springfield, USA\\
%Email: homer@thesimpsons.com}
%\IEEEauthorblockA{\IEEEauthorrefmark{3}Starfleet Academy, San Francisco, California 96678-2391\\
%Telephone: (800) 555--1212, Fax: (888) 555--1212}
%\IEEEauthorblockA{\IEEEauthorrefmark{4}Tyrell Inc., 123 Replicant Street, Los Angeles, California 90210--4321}}

% use for special paper notices
%\IEEEspecialpapernotice{(Invited Paper)}

\maketitle

\begin{abstract}
Spiking neural network (SNN) is interesting due to its strong bio-plausibility and high energy efficiency. However, its performance is falling far behind conventional deep neural networks (DNNs). In this paper, considering a general class of single-spike temporal-coded integrate-and-fire neurons, we analyze the input-output expressions of both leaky and nonleaky neurons. We show that SNNs built with leaky neurons suffer from the overly-nonlinear and overly-complex input-output response, which is the major reason for their difficult training and low performance. This reason is more fundamental than the commonly believed problem of nondifferentiable spikes. To support this claim, we show that SNNs built with nonleaky neurons can have a less-complex and less-nonlinear input-output response. They can be easily trained and can have superior performance, which is demonstrated by experimenting with the SNNs over two popular network intrusion detection datasets, i.e., the NSL-KDD and the AWID datasets. Our experiment results show that the proposed SNNs outperform a comprehensive list of DNN models and classic machine learning models. This paper demonstrates that SNNs can be promising and competitive in contrast to common beliefs.
\end{abstract}

% no keywords

% For peer review papers, you can put extra information on the cover
% page as needed:
% \ifCLASSOPTIONpeerreview
% \begin{center} \bfseries EDICS Category: 3-BBND \end{center}
% \fi
%
% For peerreview papers, this IEEEtran command inserts a page break and
% creates the second title. It will be ignored for other modes.
\IEEEpeerreviewmaketitle

%%%%%%%%%%%%%%%%%%%%%%%%%%%%%%%%%%%%%%%%%%%%%%%%%%%%%%%%%%%%%%%
\section{Introduction}

Spiking neural networks (SNNs) have great theoretical significance because of their bio-inspiration nature. They work more like biological neurons than conventional deep neural networks (DNNs). Neurons communicate via spike waveform just as biological neurons and work asynchronously, i.e., generate output spikes without waiting for all input neurons to spike. SNNs are promising practically mainly because of their strong potential of achieving high energy efficiency, which is important for mobile and IoT applications involving large and deep neural networks. The asynchronous working principle potentially leads to low latency that is attractive for high-speed applications \cite{tavanaei2019deep}.

%With the rapid development of deep neural networks (DNNs), it has become interesting to study whether the spike-based SNNs can also achieve success in practical applications in image recognition, object detection, speech recognition, etc.
%One of attractive features is that the asynchronous working principle of SNN leads to sparse spikes with extremely high energy efficiency. There are some efforts to implement SNNs in neuromorphic hardware to scoup this benefits.
%However, conventional ANNs, which use digital-valued neurons, lack the time dynamics and are less inspired by the biological neuronal processing and learning mechanism in the brain cortex \cite{kheradpisheh2018stdp}.  Inspired from this mechanism, spiking neural networks (SNNs) are proposed, in which spikes are sparse in time and the information being propagated is encoded with spike related temporal information, e.g., time latencies and spike rates \cite{gerstner2014neuronal}.
%As summarized in \cite{tavanaei2019deep}, SNNs have great potential in energy efficiency when implemented on neuromorphic architecture.

Although the study of SNN has a very long history, its development has fallen far behind DNNs. Most of the studies are still limited to shallow networks rather than deep networks. Even for shallow SNNs, the training is still very difficult, the convergence is still very slow, and the testing performance is still much worse than DNNs. One of the commonly believed reasons is that SNNs use discrete spikes that are nondifferentiable. Nevertheless, although many techniques have been developed to approximate the nondifferentiable spiking pulses with smooth and differentiable pulses, the training and performance of SNNs are still not encouraging \cite{neftci2019surrogate}. 

Another less-mentioned reason for the difficulty of training comes from the computational complexity perspective. Each SNN neuron's output is a time waveform rather than a single number as in the conventional DNN. Since the input-output response is described by a differential equation, many SNN training methods rely on solving such a differential equation online during training, which is computationally prohibitive in both the forward inference and the backward gradient back-propagation when the number of neurons becomes large. 

Nevertheless, the third reason, which we believe more fundamental than the above two reasons but has not aroused too much attention, is that the SNN neurons adopted in most existing works have overly-complex and overly-nonlinear input-output responses that lead to slow and ill convergence during training and thus low performance during testing. Experience over DNNs has shown that while sufficient nonlinearity of the whole network is necessary, simplifying every single neuron's input-output nonlinearity can enhance both training speed and convergence \cite{goodfellow2016deep}. Simplifying every single neuron's input-output expression also greatly reduces computational complexity and speeds up training, but this is secondary to the convergence issue.

Many different spiking neuron models have been used in SNNs, with leaky and nonleaky neurons, various spikes waveforms, and various kernel functions. There are a variety of different information coding methods, such as encoding information over spike rate, spike count, spike time, etc. Such variety is caused inherently by the variety of biological neurons. In general, neurons with higher bio-plausibility have more complex input-output expressions and are thus more difficult to train. SNNs should be developed as a trade-off between bio-plausibility and practicality. 

In this paper, our first focus is to analyze the input-output expressions of the two general classes of SNN neurons, i.e., leaky and nonleaky neurons with single-spike temporal coding. We will show that leaky neurons are too complex to train while nonleaky neurons can build SNNs with easy training and high performance. 

%In particular, because we use the continuous spiking time as information carrier, in software implementation, we can directly use the close-form solution of the spike waveform to derive the nonlinear layer-by-layer mapping function. As a result, direct training of multiple layer deep SNN becomes a feasbile task.

Our second focus is to demonstrate that SNNs built with nonleaky neurons can be superior to conventional DNNs once the training hurdle is overcome. We experiment with the SNNs over two popular network intrusion detection datasets: NSL-KDD and AWID \cite{lopez2020application}, and show that they outperform a comprehensive list of existing DNN and classic machine learning algorithms. 

The major contributions of this paper are listed as follows.

\begin{itemize}
    \item We analyze the input-output response of two general types of SNN neurons to show that the commonly used leaky neurons have too complex and too nonlinear input-output responses and are thus hard to train. 
    \item We show that SNNs built with nonleaky neurons can have much less complex and much less nonlinear input-output response. They can be trained as easily as conventional DNNs. We provide a new training algorithm.
    \item We train the proposed SNNs over two popular network intrusion detection datasets NSL-KDD and AWID. New benchmark results are obtained. To the best of our knowledge, this is the first time that SNN is reported for these two network intrusion detection datasets. 
    %The testing accuracy degradation is within $4\%$ when the weight resolution is reduced to just $4$-bit quantization.
\end{itemize}

This paper is organized as follows. Related works are introduced in Section \ref{sec2}. The SNN neuron's input-output expressions are analyzed in Section \ref{sec3}. The network intrusion datasets are described in Section \ref{sec4}. Experiments are presented in Section \ref{sec5} and conclusions are given in Section \ref{sec6}.

%The structure of the article is as follows. In Section \ref{sec3}, we present our proposed spiking convolutional neural networks. In Section 4, we introduce the analog circuit for energy consumption examination. In Section 5, we conduct some experiments to explore the performance of the method and provide the discussion of the simulated results. In Section 6, we draw the conclusions.

\section{Related Work} \label{sec2}

SNNs use spike pattern flows to encode spatio-temporal information \cite{wu2018spatio} \cite{kheradpisheh2018stdp}. The neurons can in general be classified into two classes: leaky neuron and nonleaky neuron. Most of the existing SNN works have focused on leaky neurons because it is considered as more bio-plausible. 

SNN training methods can be categorized into three classes: unsupervised learning,
supervised learning with indirect training, and supervised learning with direct training. For unsupervised learning, spike timing-dependent plasticity (STDP) is the most well-known one, which adjusts the weights connecting the pre- and post-synaptic neurons based on their relative spike times \cite{caporale2008spike,diehl2015unsupervised,kheradpisheh2018stdp,lee2018deep}. The dependency on the local neuronal activities without a global supervisor makes it have low performance.

For the second class, the most successful approach is to translate trained DNNs to SNNs \cite{tavanaei2019deep}\cite{rueckauer2017conversion}. Zhang et al. \cite{zhang2019tdsnn} converted VGG16 to SNN with single-spike temporal coding. Although the translation approach can get the highest accuracy performance among SNNs so far, the loss of spike sparsity degrades energy efficiency. 

For the third class, SpikeProp \cite{bohte2002error} minimized the loss between the true firing time and the single desired firing time with gradient descent rule over soft nonlinearity models. Gardner et al. \cite{gardner2015learning} applied a probability neuron model to calculate gradients. With spike rate coding, gradients were calculated after impulse spikes were approximated by smooth functions in \cite{hunsberger2015spiking,lee2016training}. Single-spike temporal coding was adopted in \cite{mostafa2017supervised} where direct training was demonstrated in shallow networks over the MNIST dataset. Wu et al. \cite{wu2019direct} conducted direct training based on spatio-temporal back-propagation, where ``spatio" referred to layer-by-layer gradient propagation and ``temporal" referred to gradient propagation through time-domain spike waveform. Neftci et al. \cite{neftci2019surrogate} used a recurrent neural network to calculate the time-domain spike waveform of nonleaky neurons.

Intrusion poses a serious risk in a network environment. Effective and efficient ways to detect the intrusion attacks and to enhance the network's ability to detect many types of intrusions have become important. Network intrusion detection algorithms can be categorized into three main groups: Statistics-based \cite{lin2015cann}, knowledge-based \cite{elhag2015combination}\cite{butun2013survey}, and machine learning-based \cite{buczak2015survey}\cite{meshram2017anomaly}. The statistics-based approach involves collecting and examining every data record in a set of items and building a statistical model of normal user behavior. The knowledge-based approach tries to identify the requested actions from existing system data such as protocol specifications and network traffic instances. In contrast, the machine-learning method acquires complex pattern-matching capabilities from training data. Niyaz et al. studied deep learning over the NSL-KDD network intrusion detection dataset \cite{javaid2016deep}. Shone et al. applied autoencoder \cite{shone2018deep} while Yin et al. applied a recurrent neural network to classify the NSL-KDD dataset \cite{yin2017deep}. Revyz et al. used deep learning to classify the AWID dataset \cite{rezvy2019efficient}. Lopez-Martin et al. \cite{lopez2020application} compared the performance of reinforcement learning against a list of other machine learning methods over both the NSL-KDD and AWID datasets \cite{lopez2020application}.

\section{Neuron Models in Spiking Neural Networks} \label{sec3}
In this section, we first analyze the input-output expressions of nonleaky and leaky integrate-and-fire spiking neurons. Then we develop SNN models based on the nonleaky neurons. Many SNNs use spike counts or rates to encode information, which may not be
energy and time efficient enough because either a long time duration or a high rate count is needed for precise information estimation. Therefore, we consider neurons that use spiking time to encode information, and each neuron emits a single spike only for energy efficiency. 

\subsection{Input-output response of nonleaky spiking neuron}

\begin{figure}[t]
\centering
\fbox{\includegraphics[width=0.45\linewidth]{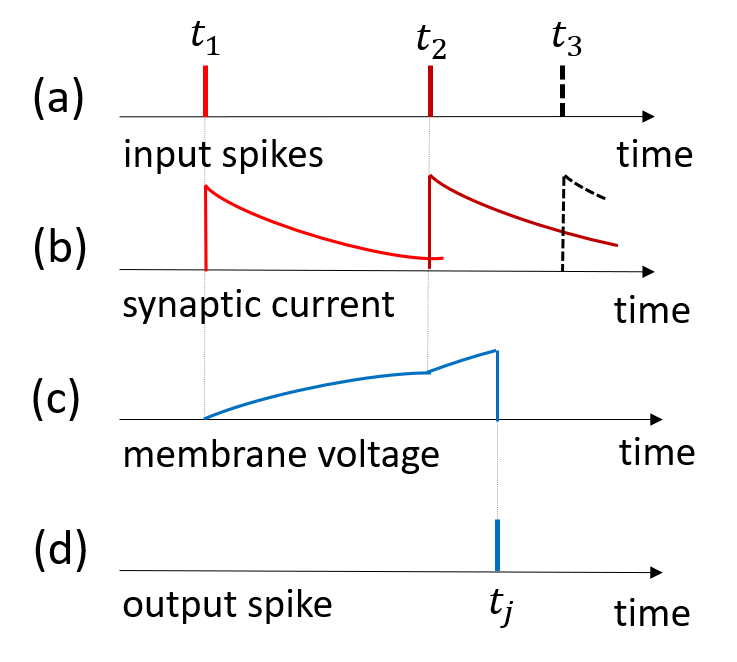}} \fbox{\includegraphics[width=0.45\linewidth]{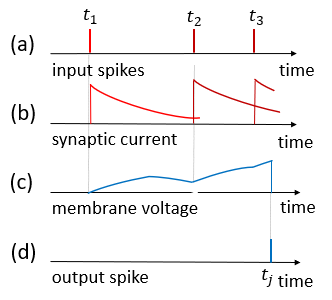}}
\centerline{(i) n-LIF neuron\hspace{0.25\linewidth} (ii) LIF neuron}
\caption{Spiking time and waveform for (i) the n-LIF neuron and (ii) the LIF neuron. (a) Three input neurons spike at time $t_1, t_2, t_3$, respectively. (b) Synaptic current $g (t-t_i)$ jumps at time $t_i$ and extends for $T$. (c) Membrane voltage potential $v_j(t)$ rises towards the firing threshold. (d) The output neuron $j$ emits a spike at time $t_j$ when the threshold is crossed.}
\label{fig:neuroncircuit}
\end{figure}

For the non-leaky integrate-and-fire (n-LIF) neuron, if spiking time is used as information carrier, the membrane potential $v_j(t)$ of the neuron $j$ is in general described by
%\begin{small}
\begin{equation}
%\centering
\frac{dv_{j}(t)}{dt} = \sum_{i} w_{ji} g(t-t_{i}), 
%\;\;\;\;  {\rm where} \;\; \kappa(t)=u(t)e^{-\frac{t}{\tau}}, 
\label{eq3.11}
\end{equation}
%\end{small}
where $w_{ji}$ is the weight of the synaptic connection from the input neuron $i$ to the output neuron $j$, $t_i$ is the spiking time of the input neuron $i$, $g(t)$ is the synaptic current kernel function or spike waveform. The value of neuron $i$ is encoded in the spike time $t_i$.  We assume $g(t)=0$ for $t<0$ or $t>T$. 
%We use exponential decaying as given below
%\begin{equation}
%\kappa(t)=u(t)e^{-\frac{t}{\tau}},
%\end{equation}
%where $\tau$ is the decaying time constant, and $u(t)$ is the unit step function defined as
%\begin{equation}
%u(t) = \left\{ \begin{array}{ll}
%1, \;\;\; & {\rm if} \;\; t\geq 0\\
%0, \;\;\; & {\rm otherwise}
%\end{array} \right. .
%\end{equation}
%\end{small}
For single-spike neuron models, a neuron is allowed to spike only once unless the network is reset or a new input pattern is presented. Fig. \ref{fig:neuroncircuit}(a) illustrates how this neuron works. 

During software implementation of the SNN or training SNN, we need to calculate $t_j$ from $t_i$. We would better avoid using numerical ODE (ordinary differential equation) solver to calculate $v_j(t)$ and $t_j$. Therefore, in this paper we focus on the ways of exploiting the solution of (\ref{eq3.11}), i.e., 
\begin{equation}
  v_j(t)=\sum_i w_{ji} G(t-t_i), \;\;\; {\rm where} \;\; G(t)=\int_{0}^{t} g(x) dx   
\end{equation}
to calculate $t_j$.
%where $u(t)$ is the unit step function, i.e., $u(t)=1$ if $t\geq 0$ and $0$ else. 
Let the spiking threshold voltage be $v_0$. Then this neuron will spike at time $t_j$ if
\begin{equation}
    v_0 = \sum_{i} w_{ji} G(t_j-t_i).  \label{eq3.20}
\end{equation}
%which can be further written as
%\begin{equation}
%    v_0 = \sum_{i \in {\cal C}} w_{ji} G(t_j-t_i), \label{eq3.20}
%\end{equation}
From (\ref{eq3.20}) we look for closed-form expressions for the output spiking time $t_j = f(t_i; w_{ji})$. Closed-form expressions can be obtained in some special cases only, e.g., when $G(t_j-t_i)$ can be separated into $G_1(t_j)G_2(t_i)$ or $G_1(t_j)+G_2(t_i)$. In the sequel, we consider two such special cases. 

\paragraph{Case 1. Unit-step spike} Let 
\begin{equation}
    g(t)=\left\{ \begin{array}{ll} a, & 0 \leq t \leq T \\
      0, & {\rm else} \end{array} \right.
\end{equation}
To simplify notation, we assume that $T > t_j$, which means the neuron spike waveform is a unit step function. Then
\begin{equation}
    G(t) = \left\{ \begin{array}{ll} at, & 0 \leq t \leq T \\
    0, & {\rm else} \end{array} \right.
\end{equation}
In this case, from (\ref{eq3.20}) we can find
\begin{equation}
    t_j = \frac{\frac{v_0}{a} + \sum_{i\in {\cal C}} w_{ji}t_i} {\sum_{i \in {\cal C}} w_{ji}},
\end{equation}
where the set ${\cal C}=\{\forall k: t_k < t_j\}$ includes all the input neurons (and only these neurons) that have spike time $t_k$ less than the output spike time $t_j$. 
The above equation can be written as the standard DNN neuron's input-output expression form (weight plus bias) as
\begin{equation}
    t_j =\sum_{i\in {\cal C}} t_i \frac{w_{ji}} {\sum_{\ell \in {\cal C}} w_{j\ell}} + \frac{v_0/a} {\sum_{\ell \in {\cal C}} w_{j\ell}}.     \label{eq3.25}
\end{equation}
The composite weights $\frac{w_{ji}} {\sum_{\ell \in {\cal C}} w_{j\ell}}$ and bias $\frac{v_0/a} {\sum_{\ell \in {\cal C}} w_{j\ell}}$ are nonlinear. There is no other nonlinear activation function. One can add extra nonlinear activation to $t_j$, but our experiments indicate that this is not necessary. Since (\ref{eq3.25}) is in a similar form to conventional DNN input-output expression, gradient back-propagation and training can be conducted in a similar way as DNN. 

\paragraph{Case 2. Exponentially-decaying spike} In this case, the spike waveform is
\begin{equation}
g(t)=\left\{ \begin{array}{ll} e^{-\frac{t}{\tau}},  & t \geq 0  \\
       0, & t < 0  \end{array} \right.  \label{eq3.25.gt}
 \end{equation}
where $\tau$ is the decaying time constant. Then we have
\begin{equation}
    G(t)=\left\{ \begin{array}{ll} \tau (1-e^{-\frac{t}{\tau}}), &  t \geq 0 \\
    0, & t<0 \end{array} \right.
\end{equation}
From (\ref{eq3.20}), the member voltage at spiking time $t_{j}$ satisfies
\begin{equation}
    v_j(t_j) = \sum_{\forall i\in {\cal C}=\{\forall k: t_k < t_{j}\}} w_{ji} {\tau} \left( 1- e^{-\frac{t_{j}-t_i}{\tau}} \right).
\end{equation}
Considering the spiking threshold $v_0$, the neuron $j$'s spike time can be calculated from
\begin{equation}
e^{\frac{t_{j}}{\tau}} = \sum_{i\in {\cal C}} e^{\frac{t_{i}}{\tau}} \frac{w_{ji}} {\sum_{\ell \in {\cal C}} w_{j\ell}-\frac{v_0}{\tau}}. \label{eq3.26}
\end{equation}
In software implementation of SNN, we can simply use $e^{t_i/\tau}$ and $e^{t_j/\tau}$ as the input and output neuron values. The input-output response (\ref{eq3.26}) is similar to the DNN neuron's input-output response. There is no bias term, and we do not need other nonlinear activation because the composite weights ${w_{ji}}/({\sum_{\ell \in {\cal C}} w_{j\ell}-v_0/\tau})$ are nonlinear.

\subsection{Input-output response of leaky spiking neuron}

For the leaky integrate-and-fire (LIF) neuron, the membrane potential of the neuron $j$ is modeled as
\begin{equation}
    \frac{dv_j(t)}{dt}+bv_j(t) = \sum_i w_{ji} g(t-t_i),  \label{eq3.51}
\end{equation}
where $b$ is a positive constant representing the decaying rate of the membrane potential when there is no input. Fig. \ref{fig:neuroncircuit}(b) illustrates how this neuron model work. Note the decaying of membrane potential after $t_1$.

To solve (\ref{eq3.51}), let $\mu(t)=e^{\int b dt} = e^{bt}$, and multiply $\mu(t)$ to both sides of (\ref{eq3.51}). We obtain
\begin{equation}
      e^{bt}\frac{dv_j(t)}{dt}+be^{bt}v_j(t) = \sum_i w_{ji} g(t-t_i)e^{bt},
\end{equation}
which can be written as
\begin{equation}
   \frac{d}{dt} \left( e^{bt}v_j(t) \right) = \sum_i w_{ji} g(t-t_i) e^{bt}. 
\end{equation}
Integrating both sides, we can get
\begin{equation}
   \int_{0}^t \frac{d}{dx} \left( e^{bx}v_j(x) \right)dx = \sum_i w_{ji} \int_{0}^t g(x-t_i) e^{bx}dx, 
\end{equation}
which can be deducted as
\begin{equation}
    v_j(t)=v_j(0)e^{-bt} + \sum_{i}w_{ji} e^{-bt} \int_0^t g(x-t_i)e^{bx}dx.
\end{equation}
Assume zero-initial condition, i.e., $v_j(0)=0$, and exploit the fact that $g(t)=0$ for $t<0$. We can get
\begin{equation}
    v_j(t_j)=\sum_{i\in {\cal C}}w_{ji} e^{-bt_j} \int_0^{t_j} g(x-t_i)e^{bx}dx. \label{eq3.60}
\end{equation}
Closed-form solutions to $t_j$ based on (\ref{eq3.60}) can only be obtained under some special cases. Note that if $b=0$, it reduces back to the nonleaky neuron case.

\paragraph{Case 1. Unit-step spike} In this case, let $g(t)=a$ for $0 \leq t \leq T$ and $0$ otherwise. Assume $T>t_j$, from (\ref{eq3.60}), at $v_j(t_j)=v_0$ we can get
\begin{align}
v_0 &=\sum_{i\in {\cal C}}w_{ji} e^{-bt_j} \int_{t_i}^{t_j} a e^{bx}dx \nonumber \\
    &= \sum_{i\in {\cal C}} w_{ji} \frac{a}{b}\left( 1-e^{b(t_i-t_j)} \right).
\end{align}
Rearranging the terms, we arrive at
\begin{equation}
    e^{bt_j} = \sum_{i\in {\cal C}} e^{bt_i} \frac{w_{ji}}{\sum_{\ell \in {\cal C}}w_{j\ell} - bv_0/a},   \label{eq3.65}
\end{equation}
which is similar to the conventional DNN input-output expressions if we let the values of the input neuron  and the output neuron be $e^{bt_i}$ and $e^{bt_j}$, respectively. Nonlinearity is embedded in the composite weights. Interestingly, this case is similar to the nonleaky with exponentially-decaying spike case, as (\ref{eq3.65}) is similar to (\ref{eq3.26}).

\paragraph{Case 2. Exponentially-decaying spike}
In this case, the spike waveform is the same as (\ref{eq3.25.gt}). From (\ref{eq3.60}) we have
\begin{align}
  v_j(t_j)&=\sum_{i\in {\cal C}} w_{ji}e^{-bt_j} \int_{t_i}^{t_j} e^{-\frac{x-t_i}{\tau}} e^{bx} dx    \nonumber \\
  &= \sum_{i\in {\cal C}}w_{ji}\frac{\tau}{1-b\tau} \left( e^{-\frac{t_j-t_i}{\tau}}-e^{-b(t_j-t_i)} \right). \label{eq3.70}
\end{align}
As pointed out in \cite{goltz2019fast}, there are only two special parameter settings that we can find closed-form solution to $t_j$. The first parameter setting is $b\tau=1$ while the second parameter setting is $b\tau=1/2$. Obviously, such settings severely limit their practical applicability.

For the first parameter setting $b\tau=1$, (\ref{eq3.70}) becomes
\begin{equation}
    \lim_{b\rightarrow \frac{1}{\tau}} v_j(t_j) = \sum_{i\in {\cal C}} w_{ji} (t_j-t_i)e^{-b(t_j-t_i)}.  \label{eq3.72}
\end{equation}
At $v_j(t_j)=v_0$, rearranging the terms of (\ref{eq3.72}) we obtain
\begin{equation}
    v_0e^{bt_j}-t_j\sum_{i\in{\cal C}}w_{ji}e^{bt_i} + \sum_{i \in {\cal C}}w_{ji}t_ie^{bt_i} =0,
\end{equation}
whose solution can be expressed as the Lambert $W$ function as
\begin{equation}
    t_j = \frac{\sum_{i\in{\cal C}}w_{ji}t_i e^{bt_i}} {\sum_{i\in{\cal C}}w_{ji}e^{bt_i}} 
    -\frac{1}{b}W\left(-\frac{bv_0}{\sum_{i\in{\cal C}}w_{ji}e^{bt_i}} e^{\frac{b\sum_{i \in {\cal C}}w_{ji}t_i e^{bt_i}} {\sum_{i\in {\cal C}}w_{ji}e^{bt_i}}}
     \right).  \label{eq3.74}
\end{equation}
Given input spiking time $t_i$ and weights $w_{ji}$, we can use (\ref{eq3.74}) to calculate output spiking time $t_j$. The gradient can be evaluated similarly since the Lambert $W$ function has closed-form gradient expression. 

\begin{figure}[t]
\centering
{\includegraphics[width=0.45\linewidth]{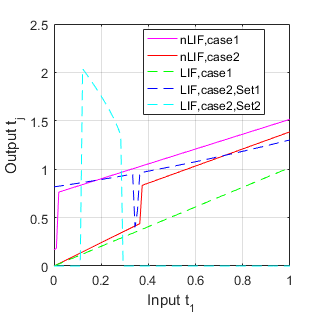}} {\includegraphics[width=0.45\linewidth]{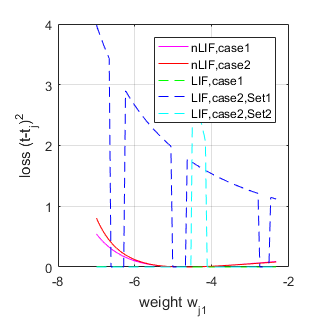}}
\centerline{(a)\hspace{0.4\linewidth} (b)}
\caption{(a) $t_j$ vs $t_i$ for $i=1$. (b) Loss function $(t-t_j)^2$ where $t_j$ is the correct output and $t$ is calculated as a function of first weight $w_{j1}$. Loss function is $0$ at the true weight. $5$ input neurons. $t_i$ and $w_{ji}$ are randomly generated, $v_0=1$, $b=0.5$, $\tau=1$, $\alpha=1$. Complex (nonreal) Lambert W function and square-root values in (\ref{eq3.74})(\ref{eq3.75}) are skipped.}
\label{fig:errorcurvers}
\end{figure}

For the second parameter setting $b\tau=1/2$, (\ref{eq3.70}) can be changed into
\begin{equation}
 \frac{v_0}{2\tau} \left( e^{\frac{t_j}{2\tau}} \right)^2 +
 \sum_{i\in{\cal C}} w_{ji}e^{\frac{t_i}{2\tau}} \left( e^{\frac{t_j}{2\tau}} \right) -
\sum_{i\in{\cal C}} w_{ji}e^{\frac{t_i}{\tau}}=0.
\end{equation}
We can find the solution as
\begin{align}
e^{\frac{t_j}{2\tau}}&=-\frac{\tau}{v_0}\sum_{i\in {\cal C}}w_{ji}e^{\frac{t_i}{2\tau}} \nonumber \\
& \pm \frac{\tau}{v_0} \sqrt{  \left(\sum_{i\in {\cal C}}w_{ji}e^{\frac{t_i}{2\tau}}\right)^2 + \frac{2v_0}{\tau} \sum_{i \in {\cal C}} w_{ji}\left( e^{\frac{t_i}{2\tau}} \right)^2 }
\label{eq3.75}
\end{align}
It can been seen that $e^{\frac{t_j}{2\tau}}$ is a function of $e^{\frac{t_i}{2\tau}}$. Therefore, we can use $e^{\frac{t_i}{2\tau}}$ and $e^{\frac{t_j}{2\tau}}$ as the input and output neuron values in the software implementation of SNN. The gradient can also be evaluated. 

Nevertheless, in both of the above two parameter settings, the computational complexity becomes prohibitive when the number of neurons becomes big. More importantly, each neuron's input-output becomes overly-complex and overly-nonlinear, which is harmful to the training of deep neural networks. Fig. \ref{fig:errorcurvers} shows some typical $t_j \sim t_i$ curves and cost function $(t-t_j)^2$ curves for the above 5 cases, from which we can see that the leaky neuron can have very complex nonlinear patterns. 

Although the above deduction is just for two special parameter settings, we expect that such overly-nonlinear property holds in general for leaky neurons and argue that this is one of the key reasons that SNNs build with leaky neurons are difficult to train and have low performance.

\subsection{SNNs Built with n-LIF Neurons}
From the above analysis we can see that it is better to adopt nonleaky neurons when building SNNs. Biological neurons are in general leaky. With periodic resetting back to zero initial state, the nonleaky neuron model can in fact approximate well the biological leaky neuron. Between the unit-step spike and the exponentially-decaying spike, we adopt the latter because it is more bio-plausible and more energy efficient. Therefore, (\ref{eq3.26}) is used to model neuron input-output response for our SNNs. 

For software implementation, we skip the details of $v_j(t)$ and just use (\ref{eq3.26}) to calculate neuron values. We also let $z_j = \alpha e^{t_j/\tau}$ with a constant $\alpha$ as the neuron value. We need to sort the input $z_i = \alpha e^{t_i/\tau}$ from small to large so as to determine the set ${\cal C}$, which leads to extra computational complexity. Note that hardware implementation does not suffer this problem because $t_i$ is the arrival timing of input spikes which are automatically sorted. 

We developed Algorithm \ref{alg1} to calculate (\ref{eq3.26}). This algorithm can be applied to all the cases studied in the previous subsection, with appropriate ${\bf z}_i$ and expressions to calculate $z_j$. For this we just need to define neuron values as
\begin{equation}
  z_i = \left\{ \begin{array}{ll} 
     \alpha t_i,  & \text{n-LIF, Case-1}  \\
     \alpha e^{t_i/\tau}, & \text{n-LIF, Case-2} \\
     \alpha e^{bt_i}, & \text{LIF, Case-1}  \\
     \alpha t_i,  & \text{LIF, Case-2, 1st Setting} \\
     \alpha e^{t_i/2\tau}, & \text{LIF, Case-2, 2nd Setting} \end{array} \right.
\end{equation}
for the various cases analyzed in the previous subsection. We also need to replace the ways of calculating ${\bf z}_{cumsum}$, ${\bf w}_{cumsum}$ and ${\bf z}_{candidate}$ of Algorithm \ref{alg1} with the corresponding equations (\ref{eq3.25}), (\ref{eq3.26}), (\ref{eq3.65}), (\ref{eq3.74}) and (\ref{eq3.75}).
%We can use the function $SORT$ and $CUMSUM$ in Tensorflow to implement the sorting and determining ${\cal C}$ conveniently, as shown in the Algorithm \ref{alg1}.

\begin{algorithm}[t]
\caption{Forward pass of a SNN neuron based on (\ref{eq3.26})}
\label{alg1}
\begin{algorithmic}
\STATE {\bf Input:} ${\bf z}=[z_1, \cdots, z_N]$: input spiking time vector
\STATE {\bf Input:} ${\bf w}=[w_1, \cdots, w_N]$: weight vector
\STATE {\bf Output:} $z_{out}$: output spiking time

\STATE ${\bf i} \gets argsort({\bf z})$: ascending order index
\STATE ${\bf z}_{sorted} \gets {\bf z}[{\bf i}]$: sorted input vector
\STATE ${\bf w}_{sorted} \gets {\bf w}[{\bf i}]$: sorted weight vector
\STATE ${\bf z}_{cumsum} \gets cumsum({\bf z}_{sorted}*{\bf w}_{sorted})$: $\sum_{i\in {\cal C}} w_{ji}z_i$
\STATE ${\bf w}_{cumsum} \gets cumsum({\bf w}_{sorted})-v_0/\tau$: $\sum_\ell w_{j\ell}-v_0/\tau$
\STATE ${\bf z}_{candidate} \gets {\bf z}_{cumsum}/{\bf w}_{cumsum}$: element-wise division
\STATE ${\bf z}_{shifted} \gets [{\bf z}_{sorted}[2:], Inf]$: next input spiking time
\STATE ${\bf c} \gets ({\bf z}_{candidate}>{\bf z}_{sorted}) \& ({\bf z}_{candidate}<={\bf z}_{shifted})$
\STATE ${k} \gets where({\bf c}==True)[1]$: index of the first True element
\STATE {\bf return} $z_{out} \gets {\bf z}_{candidate}[k]$: output spiking time
\end{algorithmic}
\end{algorithm}

The SNN model based on (\ref{eq3.26}) was initially proposed in \cite{mostafa2017supervised}. Nevertheless, only some shallow networks with $2\sim 3$ fully-connected layers were proposed and experimented over the simple XOR and MNIST dataset with mediocre performance. Compared with our Algorithm \ref{alg1}, the algorithm presented in \cite{mostafa2017supervised} does not force $t_j > t_i$ for every $i \in {\cal C}$, and thus does not match well with hardware implementation. It uses ``for" loop rather than the ``cumsum" function. In addition, it requires $\sum_{\ell \in {\cal C}} w_{j\ell}>v_0/\tau$ while we do not. This last requirement might lead to slow and worse training convergence because weights are forced to be positive biased. Since all the neuron values $z_i$ are positive, this may lead to an SNN with mostly positive weights and neuron values which may not perform well.

For an $L$-layer SNN, define the input as ${\bf z}_0$ with elements $z_{0,i} = \alpha e^{t_{0,i}/\tau}$ and the final output as ${\bf z}_{L}$ with elements $z_{L, i}=\alpha e^{t_{L, i}/\tau}$. Then we have ${\bf z}_{L} = f({\bf z}_0; {\bf w})$ with nonlinear mapping $f$ and trainable weight ${\bf w}$ which includes all weights $w_{ji}^\ell$. Let the targeting output be class $c$. We train the network with the loss function
\begin{align}
    {\cal L} ({\bf z}_L, c) &= -\ln \frac{z_{L, c}^{-1}}{\sum_{i\neq c} z_{L, i}^{-1}}
    + \lambda \sum_{\ell=1}^L \sum_{j, i} (w_{ji}^{\ell})^2
    \nonumber \\
    & + K \sum_{\ell=1}^L \sum_j \max\left\{0, \beta- \sum_i w_{ji}^\ell \right\}.   \label{eq3.80}
\end{align}
The first term is to make $z_{L, c}$ the smallest (equivalently $t_{L, c}$ the smallest) one. With time-to-first-spike (TTFS) encoding, the classification result is made at the time of the first spike among output neurons. Smaller $z_{L,c}$ or $e^{t_j}$ means stronger classification output.  The second term is $L_2$ regularization to prevent weights from becoming too large. The third term (with $\beta \geq 1$) is the weight sum cost, which enlarges each neuron's input weight summation to increase its firing probability. The parameters $K$ and $\lambda$ are weighting coefficients.

Based on the loss function (\ref{eq3.80}) and Algorithm \ref{alg1} for ${\bf z}$ calculation, we can implement the training optimization conveniently in the conventional DNN training platforms. Thanks to the closed-form expression for the neuron spiking time (\ref{eq3.26}), gradient calculation is easy and gradient-based back-propagation can be used to train the weights for our SNNs. The training becomes nothing different from conventional DNNs. 

In contrast, if we build SNNs with LIF neurons, except using the undesired unit-step spiking waveform, we do not have such easy training because the closed-form solutions to $t_j$ are too complex. 
Complex neuron input-output leads to the highly nonlinear response for each neuron, which makes the training hard to converge.
In addition, we just have two special parameter settings where the closed-form solutions exist for LIF neurons, and these two special parameter settings may not fit well with practical neuron hardware circuits. Except for these two special settings, the general LIF neurons do not even have closed-form solutions exist, which makes the computational complexity extremely high.

\section{Network Intrusion Detection Datasets}  \label{sec4}

\subsection{NSL-KDD Dataset}
The earliest effort to create an intrusion detection dataset was made by DARPA (Defense Advanced Research Project Agency) in 1998. They created the KDD98 (Knowledge Discovery and Data Mining (KDD)) dataset, which was the basis for the KDD Cup99 dataset. The KDD Cup99 dataset contained a wide variety of intrusions simulated in a military network environment. It consists of approximately 4,900,000 data records, each of which is a vector of extracted feature values from a connection record obtained from the raw network data gathered during the simulated intrusions. A connection is a sequence of TCP packets from some IP addresses. Each connection is labeled as either normal or one specific kind of attack. The simulated attacks fall into one of the following four categories:
\begin{enumerate}
    \item DOS – Denial of Service (e.g. a syn flood), 
    \item R2L – Unauthorized access from a remote machine (e.g. password guessing), 
    \item U2R – Unauthorized access to superuser or root functions (e.g. a buffer overflow attack),
    \item Probing – Surveillance and other probing for vulnerabilities (e.g. port scanning). 
\end{enumerate}

Each data record consists of 41 features, most of which take on continuous values. The features include the basic features of an individual TCP connection such as duration, protocol type, number of bytes transferred, and the flag indicating the normal or error status of the connection. Some other features of individual connections are obtained using domain knowledge, such as the number of file creation operations, the number of failed login attempts, whether root shell was obtained, and others. There are also a number of features computed using a two-second time window, such as the number of connections to the same host as the current connections within the past two seconds, percent of connections that have “SYN” and “REJ” errors, and the number of connections to the same service as the current connection within the past two seconds. 

The NSL-KDD dataset was developed from the KDD Cup99 dataset \cite{tavallaee2009detailed} after a statistical analysis performed on the KDD Cup99 dataset raised important issues that heavily influenced the intrusion detection accuracy and resulted in a misleading evaluation. Specifically, one of the main problems in the KDD Cup99 dataset is the huge amount of duplicated packets. 
Tavallaee et al. analyzed the KDD training and test datasets and revealed that approximately 78\% and 75\% of the network packets are duplicated. This huge quantity of duplicated instances in the training set would influence machine-learning methods to be biased towards normal instances and thus prevent them from learning irregular instances which are typically more damaging. 
The NSL-KDD dataset was built to resolve these problems by eliminating the duplicated records. The NSL-KDD train dataset consists of 125,973 records and the test dataset contains 22,544 records. There are 22 training intrusion attacks and 41 features. 
%among which 21 attributes refer to the connection itself and 19 attributes describe the nature of connections within the same host.

%The size of the NSL-KDD dataset is sufficient to make it practical to use the whole NSL-KDD dataset without the necessity to sample randomly. This has produced consistent and comparable results from various research works. 

In this paper, we used the KDDTrain+, KDDTest+ and KDDTest21 sets of the NSL-KDD dataset. The KDDTrain+ set contains a total of 125,973 instances comprising of 58,630 instances of attack traffics and 67,343 instances of normal traffics. The KDDTest+ set contains a total of 22,544 instances. As a subset of the KDDTest+ set, the KDDTest21 set includes a total of 11,850 instances. Cross-validation was conducted over the KDDTrain+ set in our experiments. We also considered validation using the simple hold-out (train-test) approach applied on the KDDTest+ and KDDTest-21 sets. 

%We tested the detection performance of the spiking neural network for network intrusion. 
In order to compare fairly with \cite{lopez2020application} and the results listed over there, we first used the identical datasets included with the source code of \cite{lopez2020application}. We did the same processing to extend the feature of the NSL-KDD dataset from 41 to 122. We call this the {\bf ``Original''} dataset. 

Next, to further enhance classification performance, we preprocessed the NSL-KDD dataset with a method similar to a Gaussian Receptive Field data to spike conversion, which extended the number of features of the NSL-KDD dataset from 41 to 312. We call this the {\bf ``Resampled''} dataset.  

\subsection{Aegean WiFI Intrusion Dataset (AWID)}

The AWID was published in 2015 as a collection of WiFi network data consisting of real traces of both normal and intrusion data collected from real network environments \cite{kolias2015intrusion}. Each record in the dataset is represented as a vector of 155 attributes, and each attribute has numeric or nominal values. Based on the number of target classes, the dataset can be classified into the AWID-CLS dataset and the AWIDATK dataset. The AWID-CLS dataset groups the instances into 4 main classes including normal, flooding, impersonation, and injection. The AWID-ATK dataset has 17 detailed target classes that belong to these 4 main classes. Based on the number of instances, each of these two datasets has two different versions: Full Set and Reduced Set. It is important to mention that these two versions are not related. The reduced set was collected independently from the full set at different times, with different tools, and in different environments. 

In this paper, we conducted experiments with the reduced-set version of the four-class dataset (AWID-CLS-R-Tst). The AWID-CLS-R-Tst set includes a total of 575,643 instances. To compare fairly with the results listed in \cite{lopez2020application}, first, we used the {\bf ``Original"} dataset with 46 features which were obtained from the data included in the source code of \cite{lopez2020application}. Next,
to enhance performance, we preprocessed the dataset to extend the number of features from $46$ to $206$, which we call {\bf ``Resampled"} dataset. 

%%%%%%%%%%%%%%%%%%%%%%%%%%%%%%%%%%%
\section{Experiment} \label{sec5}

\subsection{Network Models Used in Experiments}

For the NSL-KDD ``Original" dataset, we designed an SNN with three fully-connected layers: the first layer has 100 neurons, the second layer has 100 neurons, and the last layer has 5 neurons. The input is a batch vector of dimension 122. For the NSL-KDD ``Resampled" dataset, we applied this same SNN architecture, but the input is a batch vector of dimension 312.

We compared our SNN's performance with the results listed in \cite{lopez2020application}, which is fair because we used the same training and testing datasets as the latter. Nevertheless, the state-of-the-art performance listed in \cite{lopez2020application} was relatively low. As pointed out over there, much higher performance was claimed in some other literature. Unfortunately, these results claimed with high performance was not reproducible due to lack of implementation details, lack of source code, and unknown testing dataset. Such literature was omitted in \cite{lopez2020application}. Even though we also doubt these high claims, we believe that the DNN's performance may not be limited by the out-dated cases used in \cite{lopez2020application}. Therefore, we designed and trained our own DNNs, specifically, a DNN model with fully connected layers and a CNN model with 1D convolutional layers. This permits us to compare SNN more fairly with the state-of-the-art DNNs under similar experiment settings. Our DNN model has three fully-connected layers with 100, 100, and 5 neurons respectively. The architecture of the CNN is: Conv1D (312,100), Conv1D (312,100), Dropout-layer, Dense-layer (5). It has two 1D convolutional layers. Each layer has data dimension 312 and feature map dimension 100. The filter kernel size is 3. 

For the AWID ``Original" dataset, we designed an SNN with three fully-connected layers: the first layer has 100 neurons, the second layer has 100 neurons, and the last layer has 4 neurons. The input is a batch vector of dimension 46. For the AWID ``Resampled" dataset, we applied this same SNN, but the input is a batch vector of dimension 206.

Similarly, we also designed our own DNN model with fully connected layers and our own CNN model with 1D convolutional layers. This permits us to compare SNN with DNN more fairly. The DNN model has three fully-connected layers with 100, 100, and 4 neurons respectively. The architecture of the CNN model is: Conv1D (206,100), Conv1D (206,100), Dropout-layer, Dense-layer (4). The filter kernel size is 3. 

To train the SNN, we applied hyper-parameters $K=100$, $\lambda=0.001$, Adam optimizer with learning rate $0.001$ for the ``Original" dataset and $10^{-5}$ for the ``Resampled" dataset, and batch size 128. We applied $t_i = 2 d_i$ and $\alpha=1$ to map the input data $d_i$ to input neuron spiking time. We assumed $\tau=1$ and $v_0=1$ for the neurons. Through experiments, we found that keeping the training running for a lot of more iterations at a small learning rate can effectively increase classification accuracy. 

\subsection{Experiment Results over NSK-KDD Dataset}

The classification results are shown in Table \ref{tbl:kdd_testresult}. We used the following four evaluation metrics: classification accuracy, F1, precision, and recall. Note that except for our models, i.e., {\bf our DNN}, {\bf our CNN-1D}, {\bf SNN (Original)}, {\bf SNN (Resampled)}, the results of all other models were obtained from \cite{lopez2020application}. From the table, we can easily see that our CNN-1D model had classification accuracy $0.9564$ and outperformed all the results listed in \cite{lopez2020application}. More importantly, both of our SNN models outperformed all the other models, including all the DNN-based models. Our SNN achieved accuracy $0.9717$ with the original dataset, and achieved $0.9931$ accuracy with the resampled dataset. This showed that SNN can be competitive to DNNs. This also showed that appropriate data preprocessing can further enhance performance. 

\begin{table}[!t]
\caption{Compare classification results of SNN and conventional models over the NSL-KDD dataset}
\label{tbl:kdd_testresult}
\centering
\begin{tabular}{ccccc}
\hline
 &  Accuracy & F1 & Precision & Recall \\ \hline
Logistic Regression	& 0.7068	& 0.6807 &	0.8955	& 0.5491  \\ 
SVM	 
	& 0.8799 &	0.8927 &	0.9081 &	0.8779 \\ 
KNN	& 0.7808 &	0.7769	& 0.9233	& 0.6706 \\ 
Random Forest	& 0.7472	& 0.7211	& 0.9688 &	0.5743 \\ 
radient Tree Boosting	& 0.7761 &	0.7612	& 0.9690	& 0.6267  \\ 
Naïve Bayes	&	0.8019	& 0.7967 &	0.9583	& 0.6818 \\ 
AdaBoost &	0.7606	& 0.7403	& 0.9583 &	0.5992 \\
Neural Network	& 0.7966	& 0.7881 &	0.9679 &	0.6647 \\ 
CNN-1D &	0.7875	& 0.7633 &	0.8094	& 0.7875 \\ 
Reinforcement Learn	
	&0.8978 &	0.9120 &	0.8944	& 0.9303 \\ \hline \hline
Our DNN & 0.8834 & 0.8860 & 0.8936 & 0.8834 \\
Our CNN-1D & 0.9564 & 0.9561 & 0.9565 & 0.9564 \\ \hline
SNN (Original) &	0.9717	& 0.9718 &	0.9721 &	0.9717 \\
%(original) & & & & \\
SNN (Resampled)	& {\bf 0.9931}	& {\bf 0.9931}	& {\bf 0.9931}	& {\bf 0.9931} \\
%(Resampled) & & & & \\
\hline
\end{tabular}
\end{table}

Table \ref{tbl:kdd2_accuracy} shows the classification performance for each of the $5$ classes. We can see that our SNN model can classify each sample with extremely high accuracy.

\begin{table}[!t]
\caption{Itemized classification results of SNN for each of the 5 classes of the NSL-KDD ``Resampled" dataset}
\label{tbl:kdd2_accuracy}
\centering
\begin{tabular}{cccccc}
\hline
Class &	Accuracy &	F1	& Precision &	Recall & Total Data \\ \hline
normal	& 0.999379	& 0.99906	& 0.999329	& 0.998792	& 7451  \\
Dos	& 0.999334	& 0.996891	& 1	& 0.993802	& 2420 \\
R2L	& 0.999689	& 0.982801	& 0.966184	& 1	& 2754  \\
Probe	& 0.994452	& 0.97726	& 0.97922 &	0.975309	& 9705  \\
U2R	& 0.993387	& 0.992333	& 0.99116	& 0.993509	& 200 \\
\hline
\end{tabular}
\end{table}

\subsection{Experiment Results over AWID Dataset}

The classification results are shown in Table \ref{tbl:awid_testresult}. Except for our 4 models, the results of all the other models were obtained from \cite{lopez2020application}. We can see that both our DNN and our CNN-1D models achieved competitive performance as existing models. More importantly, our SNN models outperformed all the other models, including all the DNN-based models. Our SNN achieved accuracy $0.9898$ with the original dataset, and achieved $0.9984$ accuracy with the resampled dataset. This again demonstrated that the SNN can be competitive to DNNs.

\begin{table}[!t]
\caption{Compare classification results of SNN and conventional models over the AWID dataset}
\label{tbl:awid_testresult}
\centering
\begin{tabular}{ccccc}
\hline
 &  Accuracy & F1 & Precision & Recall \\ \hline
AdaBoost	& 0.9220 &	0.8850 &	0.8500 &	0.9220 \\ 
Decision Tree &	0.9620 &	0.9480 &	0.9620 &	0.9630 \\
Naïve Bayes	& 0.9055 &	0.9090 &	0.9170 &	0.9060  \\
Frequency Tabel &	0.9457 &	0.9220 &	0.9000 &	0.9460 \\ 
Random Forest &	0.9582 &	0.9440 &	0.9590 &	0.9580 \\ 
Neural Network & 0.9470 &	0.9256 &	0.9174 &	0.9473 \\ 
Reinforcement Learn &	0.9570 &	0.9394 &	0.9235 &	0.9570 \\  \hline \hline
Our DNN  & 0.9585 & 0.9624 & 0.9715 & 0.9585 \\
Our CNN-1D & 0.9528 & 0.9351 & 0.9504 & 0.9528 \\ \hline
SNN (Original)	& 0.9898 &	0.9893 &	0.9895 &	0.9898 \\
SNN (Resampled) &	{\bf 0.9984} &	{\bf 0.9985} &	{\bf 0.9985} &	{\bf 0.9984} \\ 
\hline
\end{tabular}
\end{table}

Table \ref{tbl:awid2_accuracy} shows the classification performance for each of the $4$ classes. We can see that our SNN model can classify each sample with extremely high accuracy.

\begin{table}[!t]
\caption{Itemized classification results of SNN for each of the 5 classes of the AWID ``Resampled" dataset}
\label{tbl:awid2_accuracy}
\centering
\begin{tabular}{cccccc}
\hline
Class &	Accuracy &	F1	& Precision &	Recall &	Total \\ \hline
normal &	0.999147 &	0.970569 &	0.94293 &	0.999876 &	8097 \\
flooding &	0.99931 &	0.990184 &	0.983207 &	0.997261 &	20079 \\
injection &	0.99997	 & 0.999491	& 0.999401 &	0.99958	& 16682 \\
impersonate &	0.998452 &	0.999916 &	0.999894 &	0.998427 &	530772 \\
\hline
\end{tabular}
\end{table}

\section{Conclusions} \label{sec6}
In this paper we develop single-spike temporal-coded SNNs that can be easily trained with competitive performance as conventional DNNs. We analyzed systematically the input-output expressions of single-spike temporal-coded leaky and nonleaky neurons. We show that the commonly used leaky neurons have overly-complex and overly-nonlinear responses and argue that this is the primary reason that makes SNNs hard to train and low in performance. We also show that with nonleaky neurons we can resolve these problems. We demonstrate this by experimenting with such SNNs over the two popular network intrusion detection datasets and by showing that the SNNs outperformed a list of existing methods including the DNN-based methods. The easy training and high performance indicate that SNNs can be competitive to DNNs and are promising for practical applications.

% trigger a \newpage just before the given reference
% number - used to balance the columns on the last page
% adjust value as needed - may need to be readjusted if
% the document is modified later
%\IEEEtriggeratref{8}
% The "triggered" command can be changed if desired:
%\IEEEtriggercmd{\enlargethispage{-5in}}

% references section

% can use a bibliography generated by BibTeX as a .bbl file
% BibTeX documentation can be easily obtained at:
% http://mirror.ctan.org/biblio/bibtex/contrib/doc/
% The IEEEtran BibTeX style support page is at:
% http://www.michaelshell.org/tex/ieeetran/bibtex/
%\bibliographystyle{IEEEtran}
% argument is your BibTeX string definitions and bibliography database(s)
%\bibliography{IEEEabrv,../bib/paper}
%
% <OR> manually copy in the resultant .bbl file
% set second argument of \begin to the number of references
% (used to reserve space for the reference number labels box)

\bibliographystyle{IEEEtran}
\bibliography{IEEEabrv,mybib,mybib2}

%%%%%%%%%%%%%%%%%%%%%%%%%%%%%%%%%%%%%%%%%%%%%%%%%%%%%%%%%%%%%%
%%%%%%%%%%%%%%%%%%%%%%%%%%%%%%%%%%%%%%%%%%%%%%%%%%%%%%%%%%%%%%%%
\newpage 
%$\ \ $
%\newpage
\centerline{\Large{\bf Supplementary Material}}

%\appendix

\section{Extra Experiment Results: NSL-KDD Dataset}

\begin{figure}[htbp]
\centering
\includegraphics[width=1\linewidth]{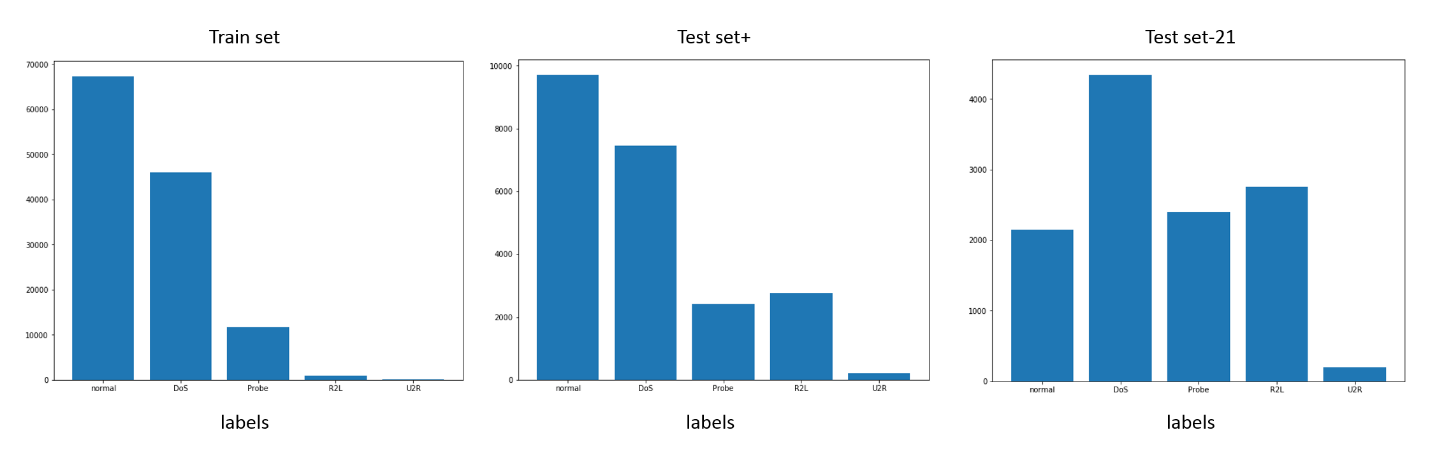}
\caption{Attack distribution in the NSL-KDD Dataset.}
\label{fig:kdd_dataset}
\end{figure}

Distribution of the five classes (normal + 4 attack types) of the NSL-KDD dataset is shown in Fig. \ref{fig:kdd_dataset}. 

\begin{figure}[htbp]
\centering
\includegraphics[width=0.9\linewidth]{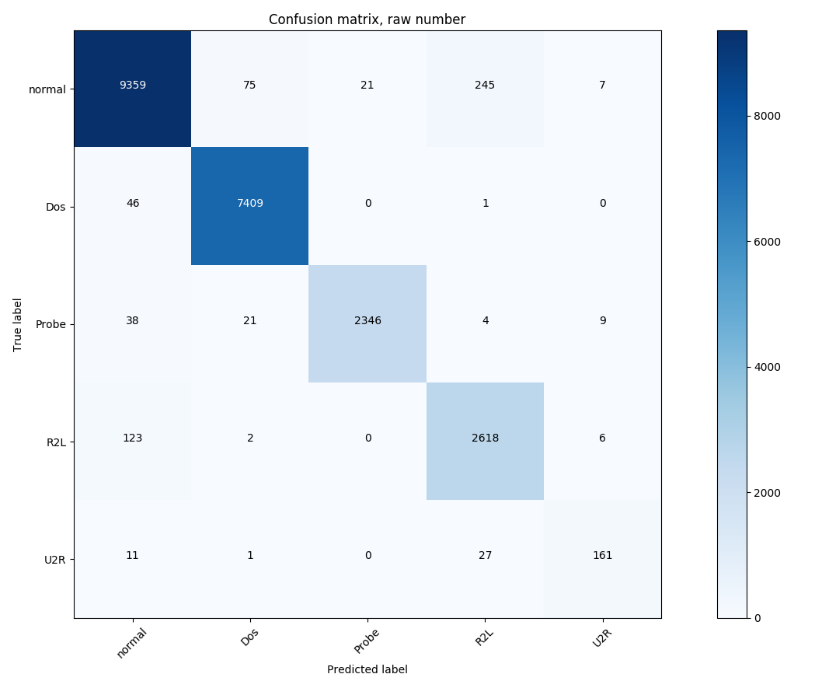}
\caption{NSL-KDD ``Original" dataset: Classification confusion matrix expressed as number of data records.}
\label{fig:kdd_rawnumber}
\end{figure}

\begin{figure}[htbp]
\centering
\includegraphics[width=0.9\linewidth]{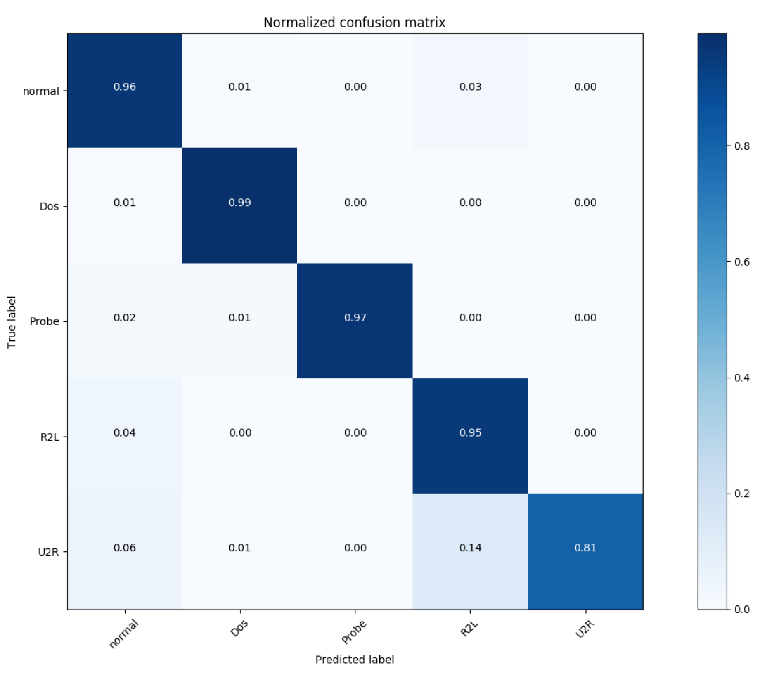}
\caption{NSL-KDD ``Original" dataset: Confusion matrix in percentage.}
\label{fig:kdd_confusionmatrix}
\end{figure}

\begin{figure}[htbp]
\centering
\includegraphics[width=0.9\linewidth]{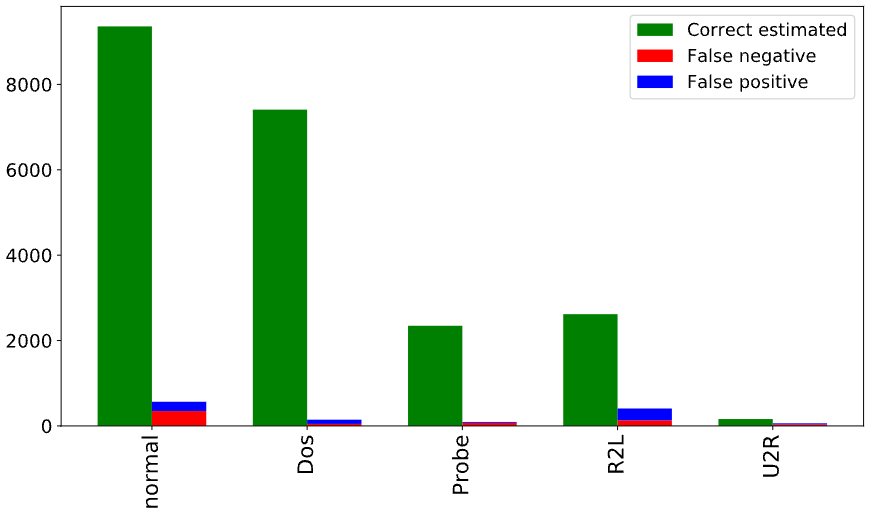}
\caption{NSL-KDD ``Original" dataset: Test result for each class.}
\label{fig:kdd_testresult}
\end{figure}

For the NSL-KDD ``Original" dataset, Table \ref{tbl:kdd_correctsamples} lists the total number of data records and the number of data records classified correctly for each class. Fig. \ref{fig:kdd_rawnumber} shows the classification confusion matrix expressed as the number of data records tested. Table \ref{tbl:kdd_accuracy} shows the classification performance for each of the 5 classes and Fig. \ref{fig:kdd_confusionmatrix} shows the confusion matrix expressed as a percentage. Fig. \ref{fig:kdd_testresult} shows the classification results for each class.

\begin{table}[htbp]
\caption{NSL-KDD ``Original" dataset: Classification results of each class. Total data records correctly classified: 21893. Total number of data records: 22530. Classification Accuracy = 97.17\%}
\label{tbl:kdd_correctsamples}
\centering
\begin{tabular}{ccccc}
\hline
 & Estimated & Correct & Total & F1 score \\
\hline
normal &	9577 &	9359 &	9707 &	97.0649  \\
Dos	& 7508	& 7409	& 7456 &	99.0243 \\
Probe	& 2367 & 	2346 &	2418 &	98.0564 \\
R2L	& 2895	& 2618 &	2749 &	92.7711 \\
U2R	& 183	& 161	& 200 &	84.0731 \\
\hline
\end{tabular}
\end{table}

\begin{table}[htbp]
\caption{NSL-KDD ``Original" dataset: Classification results of each class in percentage. Accuracy = 0.9717, F1 = 0.9718, Precision score = 0.9721, recall score = 0.9717}
\label{tbl:kdd_accuracy}
\centering
\begin{tabular}{cccccc}
\hline
Name &	Accuracy &	F1	& Precision &	Recall &	Total \\ \hline
normal	& 0.974878	& 0.970649	& 0.977237	& 0.96415	& 9707  \\
Dos	& 0.99352	& 0.990243	& 0.986814	& 0.993696	& 7456  \\
R2L	& 0.981891	& 0.927711	& 0.904318	& 0.952346	& 2418  \\
Probe &	0.995872 &	0.980564 &	0.991128 &	0.970223 &	2749 \\
U2R	& 0.997292	& 0.840731	& 0.879781	& 0.805	& 200 \\
\hline
\end{tabular}
\end{table}

\begin{figure}[htbp]
\centering
\includegraphics[width=0.9\linewidth]{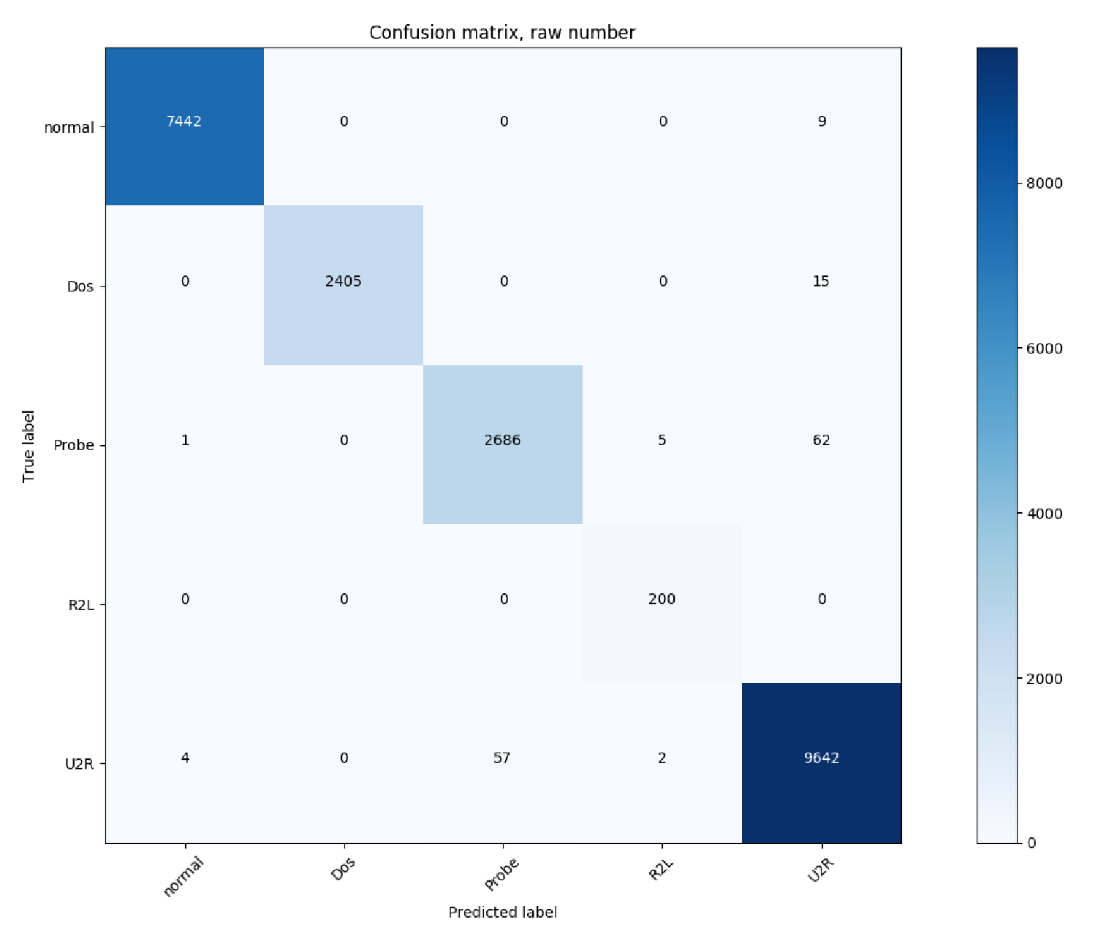}
\caption{NSL-KDD ``Resampled" dataset: Classification confusion matrix expressed as number of data records.}
\label{fig:kdd2_rawnumber}
\end{figure}

\begin{figure}[htbp]
\centering
\includegraphics[width=0.9\linewidth]{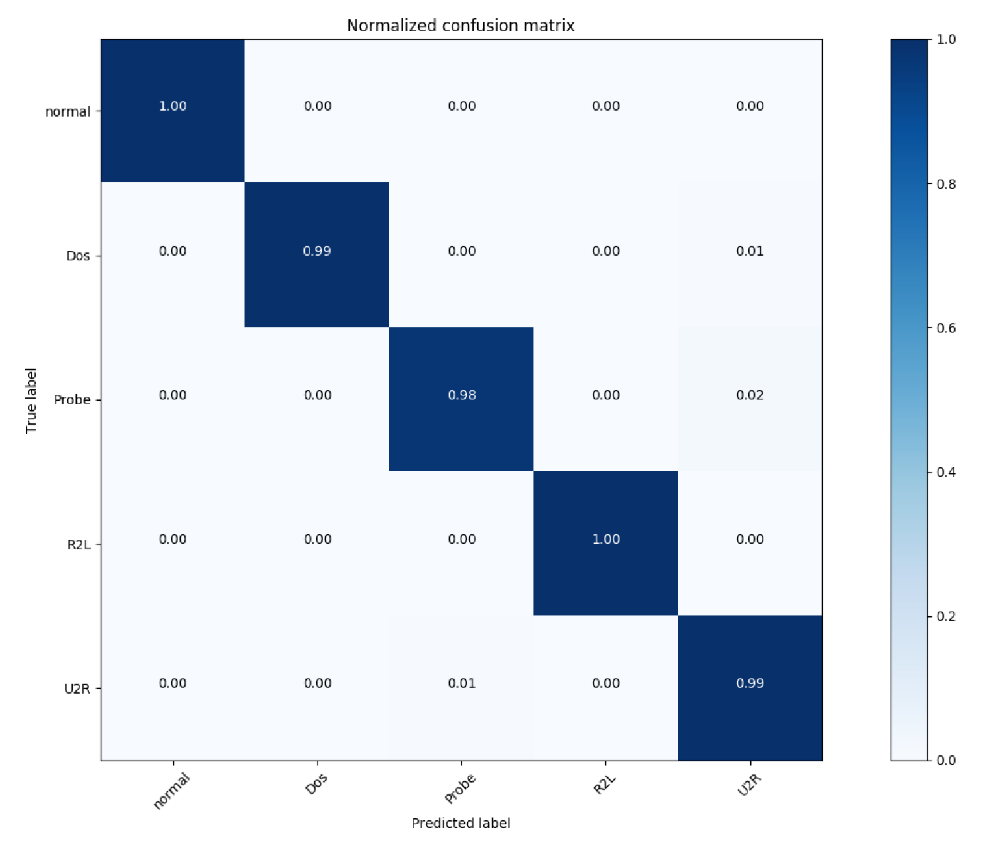}
\caption{NSL-KDD ``Resampled" dataset: Confusion matrix in percentage.}
\label{fig:kdd2_confusionmatrix}
\end{figure}

\begin{figure}[htbp]
\centering
\includegraphics[width=0.9\linewidth]{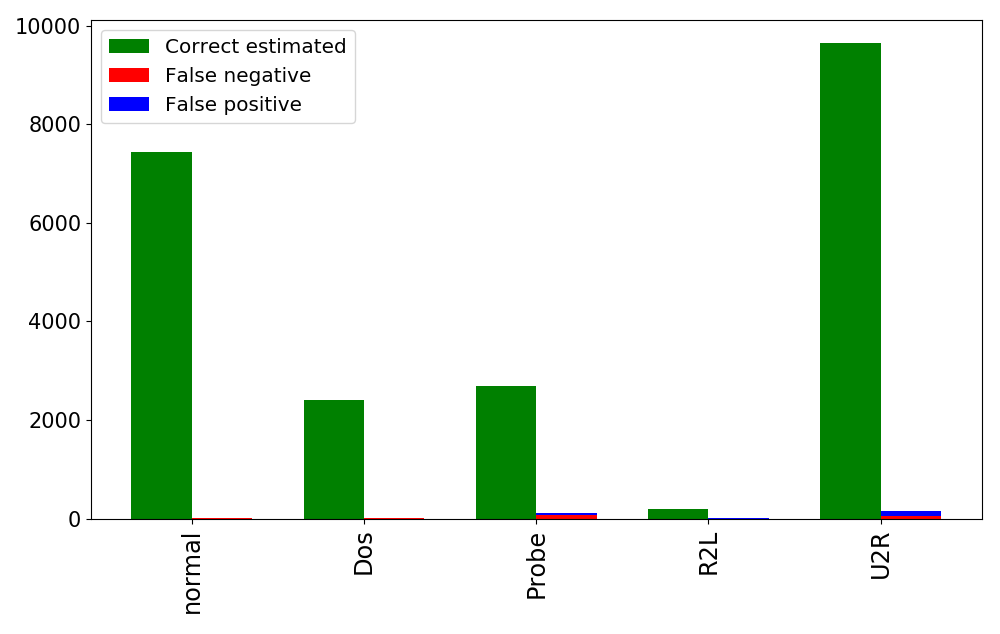}
\caption{NSL-KDD ``Resampled" dataset: Test result for each class.}
\label{fig:kdd2_testresult}
\end{figure}

For the NSL-KDD ``Resampled" dataset, Table \ref{tbl:kdd2_correctsamples} lists the total number of data records and the number of data records classified correctly for each class. Fig. \ref{fig:kdd2_rawnumber} shows the classification confusion matrix expressed as the number of data records tested. Table \ref{tbl:kdd2_accuracy} shows the classification performance for each of the 5 classes and Fig. \ref{fig:kdd2_confusionmatrix} shows the confusion matrix expressed as percentages. Fig. \ref{fig:kdd2_testresult} shows the classification results for each class.

\begin{table}[htbp]
\caption{NSL-KDD ``Resampled" dataset: Classification results of each class. Total data records correctly classified: 22375. Total number of data records: 22530. Classification Accuracy = 99.31\%}
\label{tbl:kdd2_correctsamples}
\centering
\begin{tabular}{ccccc}
\hline
 &	Estimated	& Correct	& Total	& F1 score  \\ \hline
normal	& 7447	& 7442	& 7451	& 99.906  \\
Dos	& 2405	& 2405	& 2420	&  99.6891  \\
Probe	& 2743	& 2686	& 2754	& 97.726 \\
R2L	& 207	& 200	& 200	& 98.2801  \\
U2R	& 9728	& 9642	& 9705	& 99.2333 \\
\hline
\end{tabular}
\end{table}

\section{Extra Experiment Results: AWID Dataset}

\begin{figure}[htbp]
\centering
\includegraphics[width=1\linewidth]{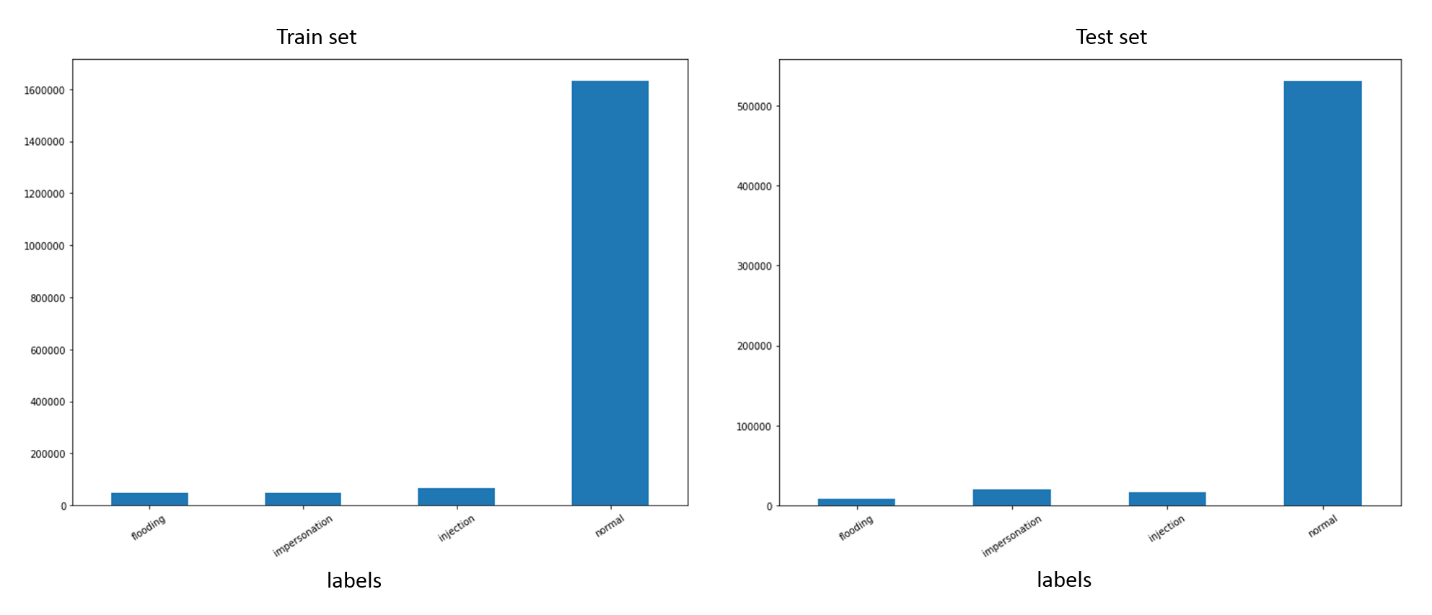}
\caption{Attack distribution in the AWID Dataset.}
\label{fig:awid_dataset}
\end{figure}

The distribution of 4 classes (normal + 3 attack classes) of the AWID dataset is shown in Fig. \ref{fig:awid_dataset}.

\begin{figure}[htbp]
\centering
\includegraphics[width=0.9\linewidth]{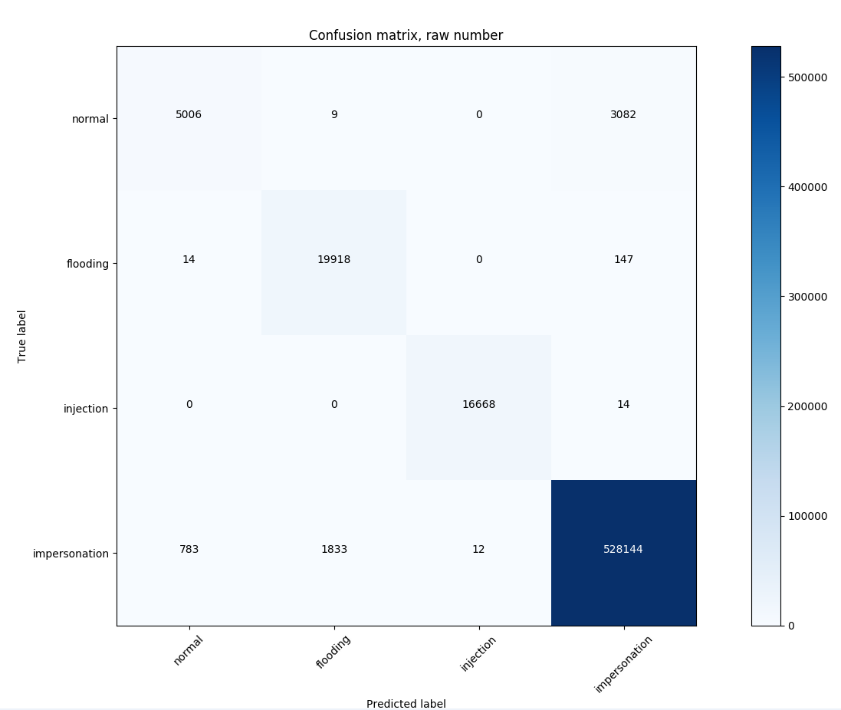}
\caption{AWID ``Original" dataset: Classification confusion matrix expressed as number of data records.}
\label{fig:awid_rawnumber}
\end{figure}

\begin{figure}[htbp]
\centering
\includegraphics[width=0.9\linewidth]{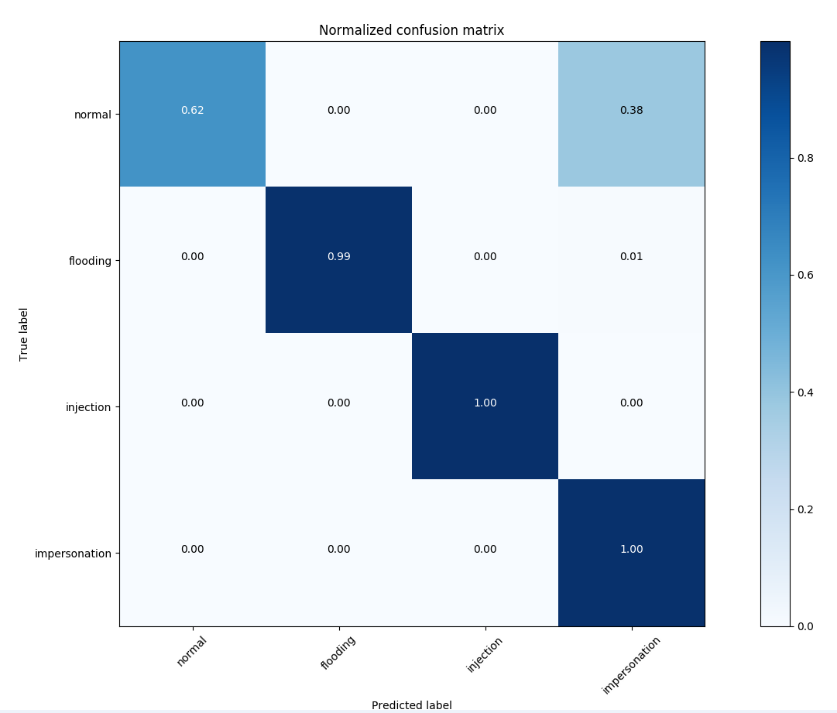}
\caption{AWID ``Original" dataset: Confusion matrix in percentage.}
\label{fig:awid_confusionmatrix}
\end{figure}

\begin{figure}[htbp]
\centering
\includegraphics[width=0.9\linewidth]{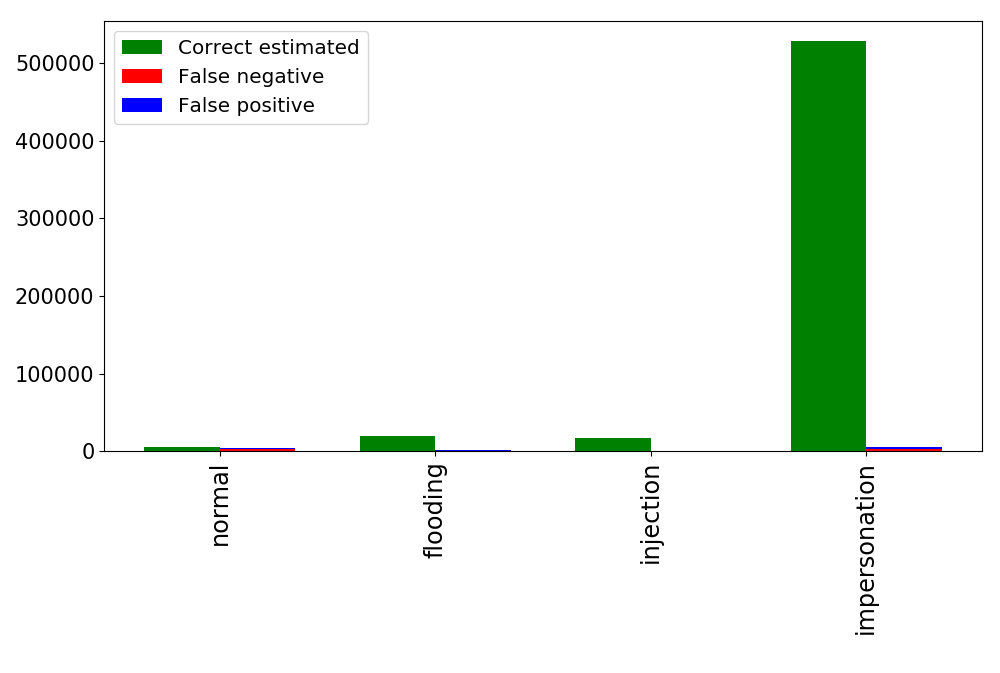}
\caption{AWID ``Original" dataset: Test result for each class.}
\label{fig:awid_testresult}
\end{figure}

For the AWID ``Original" dataset, Table \ref{tbl:awid_correctsamples} lists the total number of data records and the number of data records classified correctly for each class. Fig. \ref{fig:awid_rawnumber} shows the classification confusion matrix expressed as the number of data records tested. Table \ref{tbl:awid_accuracy} shows the classification performance for each of the 5 classes and Fig. \ref{fig:awid_confusionmatrix} shows the confusion matrix expressed as percentages. Fig. \ref{fig:awid_testresult} shows the classification results for each class.

\begin{table}[htbp]
\caption{AWID ``Original" dataset: Classification results of each class. Total data records correctly classified: 569,736. Total number of data records: 575,630. Classification Accuracy = 98.98\%}
\label{tbl:awid_correctsamples}
\centering
\begin{tabular}{ccccc}
\hline
 &	Estimated	& Correct	& Total	& F1 score  \\ \hline
normal	& 5803 &	5006 &	8097 &	72.0288 \\
flooding &	21760 &	19918 &	20079 &	95.2126 \\
injection &	16680 &	16668 &	16682 &	99.9221 \\
impersonation &	531387 &	528144 &	530772 &	99.4473 \\
\hline
\end{tabular}
\end{table}

\begin{table}[htbp]
\caption{AWID ``Original" dataset: Classification results of each class in percentage. 
 Accuracy = 0.9898, F1 = 0.9893, Precision score = 0.9895, recall score = 0.9898}
\label{tbl:awid_accuracy}
\centering
\begin{tabular}{cccccc}
\hline
Name &	Accuracy &	F1	& Precision &	Recall &	Total \\ \hline
normal & 0.993246 &	0.720288 &	0.862657 &	0.618254 &	8097 \\
flooding &	0.99652 &	0.952126 &	0.915349 &	0.991982 &	20079 \\
injection &	0.999955 &	0.999221 &	0.999281 &	0.999161 &	16682 \\
impersonation &	0.989801 &	0.994473 &	0.993897 &	0.995049 &	530772 \\
\hline
\end{tabular}
\end{table}

\begin{figure}[htbp]
\centering
\includegraphics[width=0.9\linewidth]{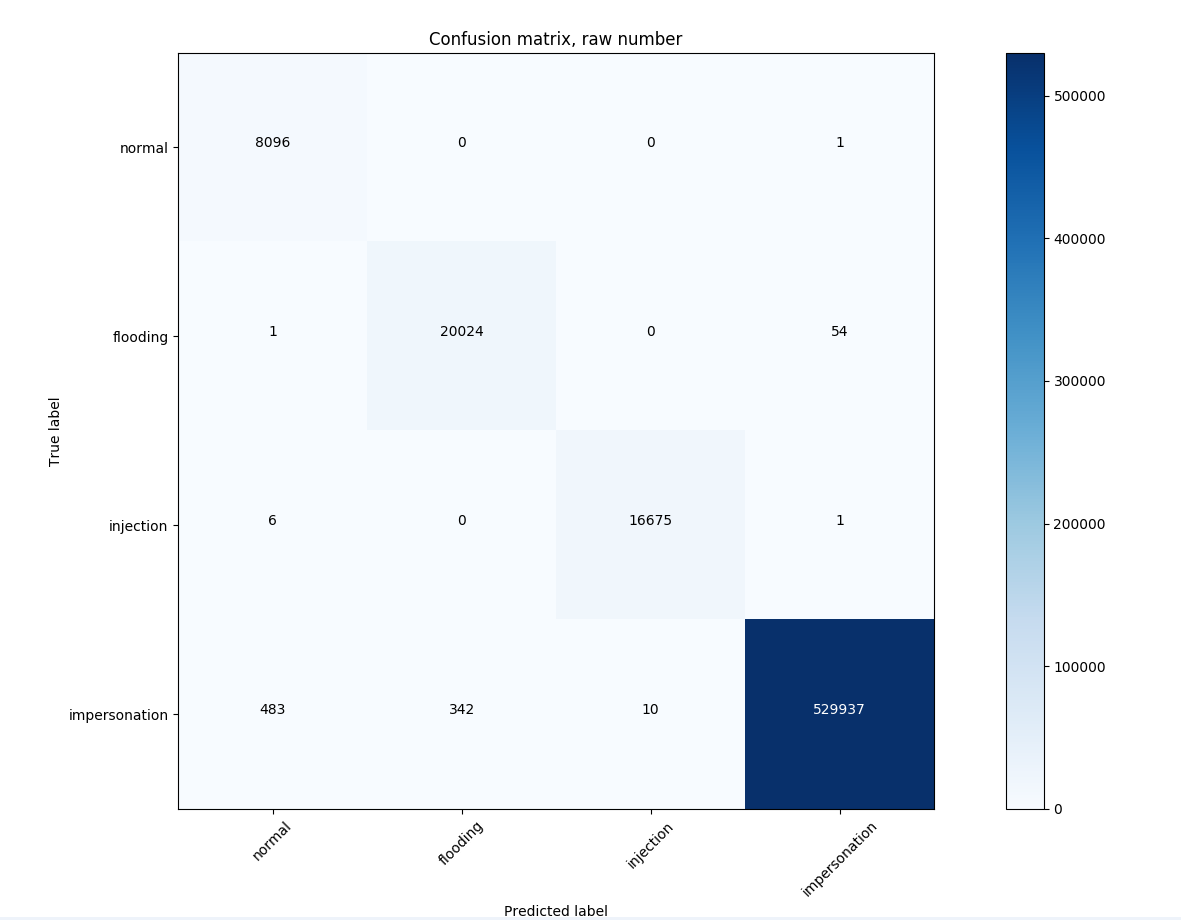}
\caption{AWID ``Resampled" dataset: Classification confusion matrix expressed as number of data records.}
\label{fig:awid2_rawnumber}
\end{figure}

\begin{figure}[htbp]
\centering
\includegraphics[width=0.9\linewidth]{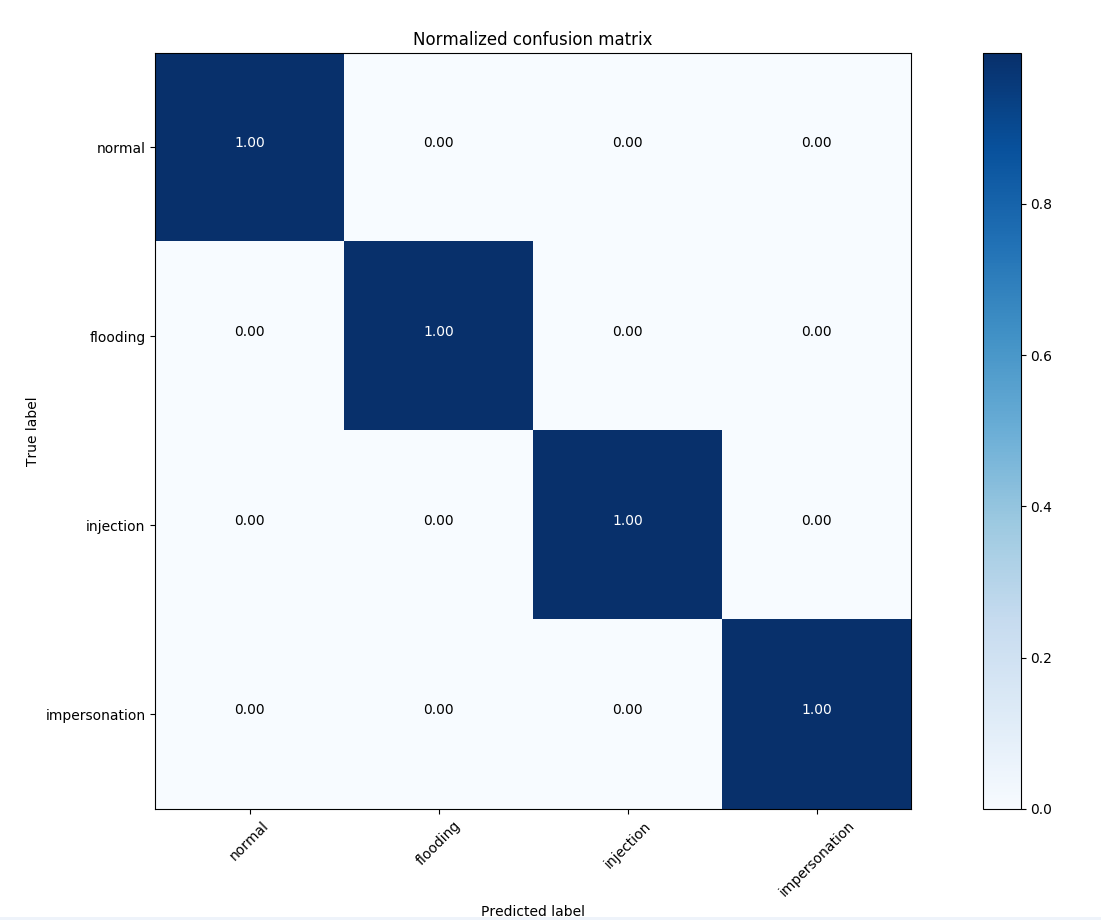}
\caption{AWID ``Resampled" dataset: Confusion matrix in percentage.}
\label{fig:awid2_confusionmatrix}
\end{figure}

\begin{figure}[htbp]
\centering
\includegraphics[width=0.9\linewidth]{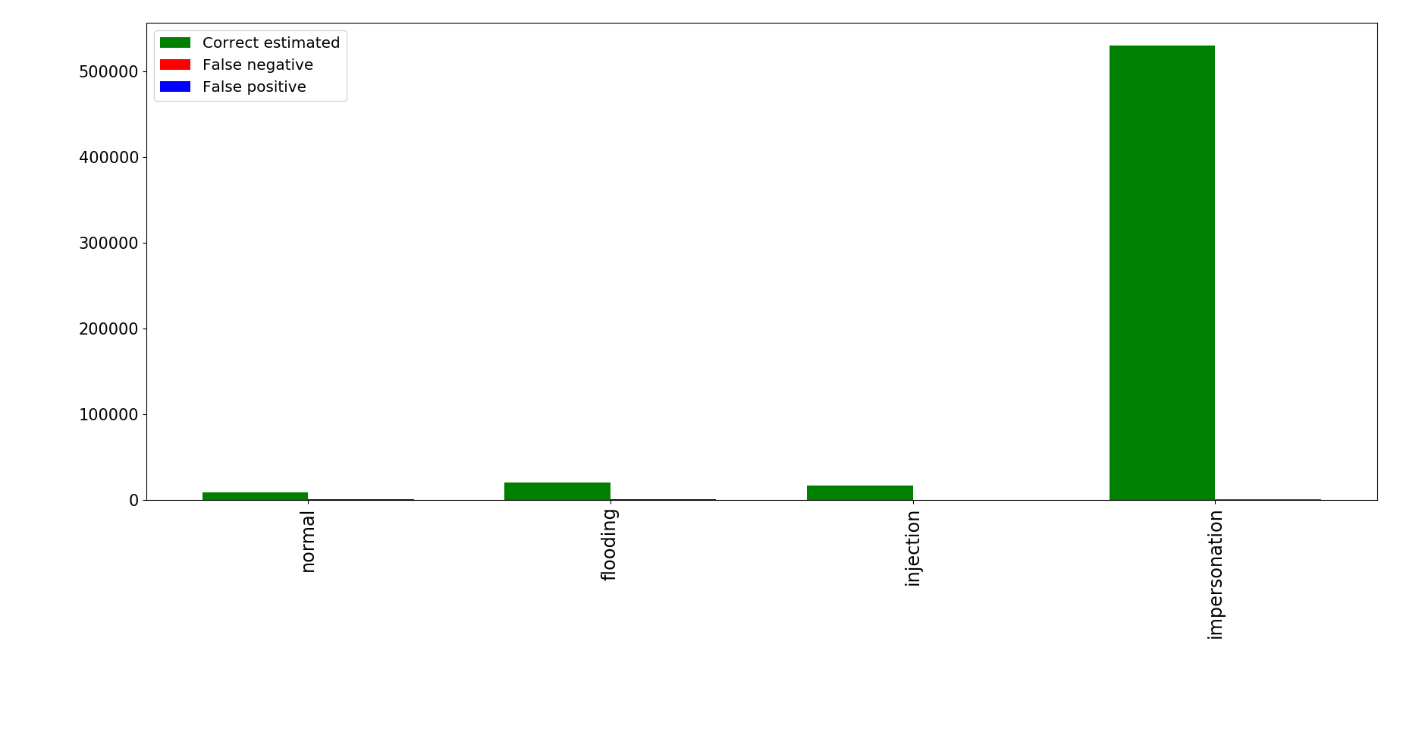}
\caption{AWID ``Resampled" dataset: Test result for each class.}
\label{fig:awid2_testresult}
\end{figure}

For the AWID ``Resampled" dataset, Table \ref{tbl:awid2_correctsamples} lists the total number of data records and the number of data records classified correctly for each class. Fig. \ref{fig:awid2_rawnumber} shows the classification confusion matrix expressed as the number of data records tested. Table \ref{tbl:awid2_accuracy} shows the classification performance for each of the 5 classes and Fig. \ref{fig:awid2_confusionmatrix} shows the confusion matrix expressed as percentages. Fig. \ref{fig:awid2_testresult} shows the classification results for each class.

\begin{table}[htbp]
\caption{AWID ``Resampled" dataset: Classification results of each class. Total data records correctly classified: 574,732. Total number of data records: 575,630. Classification Accuracy = 99.84\%}
\label{tbl:awid2_correctsamples}
\centering
\begin{tabular}{ccccc}
\hline
 &	Estimated	& Correct	& Total	& F1 score  \\ \hline
normal	& 8586 &	8096 &	8097 &	97.0569 \\
flooding &	20366 &	20024 &	20079 &	99.0184 \\
injection &	16685 &	16675 &	16682 &	99.9491 \\
impersonation &	529993 &	529937 &	530772 &	99.916 \\
\hline
\end{tabular}
\end{table}

\section{Extra Experiment Results of DNN}

For our DNN model, its performance over the NSL-KDD dataset on each attack class is listed in Table \ref{tbl:dnnkdd_correctsamples} and Table \ref{tbl:dnnkdd_accuracy}. Its performance over the AWID dataset on each attack class is listed in Table \ref{tbl:dnnawid_correctsamples} and Table \ref{tbl:dnnawid_accuracy}.

\begin{table}[htbp]
\caption{DNN over NSL-KDD dataset: Classification results of each class. Total data records correctly classified: 19915. Total number of data records: 22544. Classification Accuracy = 88.34\%}
\label{tbl:dnnkdd_correctsamples}
\centering
\begin{tabular}{ccccc}
\hline
 &	Estimated	& Correct	& Total	& F1 score  \\ \hline
 normal &       6818    &      6528 &       7458 &      91.4542 \\
Dos      &       2488   &       1935    &    2421 &     78.8348 \\
Probe &          2301   &       2107    &    2754 &     83.363  \\
R2L &            444    &       103     &      200 &      31.9876 \\
U2R &           10493  &      9242  &      9711     & 91.4868 \\
\hline
\end{tabular}
\end{table}

\begin{table}[htbp]
\caption{DNN over NSL-KDD dataset: Classification results of each class in percentage. 
 Accuracy = 0.8834, F1 = 0.8860, Precision score = 0.8936, recall score = 0.8834}
\label{tbl:dnnkdd_accuracy}
\centering
\begin{tabular}{ccccc}
\hline
Name &	Accuracy &	F1	& Precision &	Recall \\ \hline
normal & 0.945884 & 0.914542 &  0.957466 &  0.875302 \\
DoS & 0.953912 &  0.788348 &  0.777733 &  0.799257  \\
Probe & 0.962695 &  0.83363     &0.915689 &  0.765069 \\
R2L & 0.980571 &  0.319876 &  0.231982 &    0.515 \\
U2R & 0.923705 &  0.914868 &  0.880778  & 0.951704 \\
\hline
\end{tabular}
\end{table}

\begin{table}[htbp]
\caption{DNN over AWID dataset: Classification results of each class. Total data records correctly classified: 551,743. Total number of data records: 575,643. Classification Accuracy = 95.85\%}
\label{tbl:dnnawid_correctsamples}
\centering
\begin{tabular}{ccccc}
\hline
 &	Estimated	& Correct	& Total	& F1 score  \\ \hline
normal        &         7843    &      5055 &      8097 &       63.4253  \\
flooding        &      36782    &     19982 &    20079 &     70.2837 \\
injection  &           17923    &     16681 &    16682 &      96.408 \\
impersonation &   513095    &   510025 &   530785 &   97.7172 \\
\hline
\end{tabular}
\end{table}

\begin{table}[htbp]
\caption{DNN over AWID dataset: Classification results of each class in percentage. 
 Accuracy = 0.9585, F1 = 0.9624, Precision score = 0.9715, recall score = 0.9585}
\label{tbl:dnnawid_accuracy}
\centering
\begin{tabular}{ccccc}
\hline
Name &	Accuracy &	F1	& Precision &	Recall \\ \hline
 normal & 0.989872  &     0.634253 &   0.644524 &    0.624305 \\
 flooding & 0.970647    &  0.702837 &    0.543255 &   0.995169 \\
  injection & 0.997841  &     0.96408 &     0.930704 &     0.99994 \\
 impersonation &  0.958603 & 0.977172 &   0.994017  &   0.960888 \\
\hline
\end{tabular}
\end{table}

\section{Extra Experiment Results of CNN-1D}

For our CNN-1D model, its performance over the NSL-KDD dataset on each attack class is listed in Table \ref{tbl:cnnkdd_correctsamples} and Table \ref{tbl:cnnkdd_accuracy}. Its performance over the AWID dataset on each attack class is listed in Table \ref{tbl:cnnawid_correctsamples} and Table \ref{tbl:cnnawid_accuracy}.

\begin{table}[htbp]
\caption{CNN-1D over NSL-KDD dataset: Classification results of each class. Total data records correctly classified: 21561. Total number of data records: 22544. Classification Accuracy = 95.64\%}
\label{tbl:cnnkdd_correctsamples}
\centering
\begin{tabular}{ccccc}
\hline
 &	Estimated	& Correct	& Total	& F1 score  \\ \hline
normal   &     7430     &         7350 &       7458 &        98.7372 \\
Dos     &        2621  &            2385    &    2421 &        94.6053 \\
Probe  &        2638    &          2393 &       2754    &    88.7611  \\
R2L     &       159     &           143     &     200  &        79.6657  \\
U2R     &       9696    &         9290 &        9711    &    95.7387 \\
\hline
\end{tabular}
\end{table}

\begin{table}[htbp]
\caption{CNN-1D over NSL-KDD dataset: Classification results of each class in percentage. 
 Accuracy = 0.9564, F1 = 0.9561, Precision score = 0.9565, recall score = 0.9564}
\label{tbl:cnnkdd_accuracy}
\centering
\begin{tabular}{ccccc}
\hline
Name &	Accuracy &	F1	& Precision &	Recall \\ \hline
normal & 0.991661 & 0.987372 & 0.989233 & 0.985519  \\
DoS & 0.987935 & 0.946053 & 0.909958  & 0.98513  \\
Probe & 0.973119 & 0.887611 & 0.907127 & 0.868918  \\
 R2L & 0.996762 & 0.796657 & 0.899371  &   0.715  \\
  U2R & 0.963316  & 0.957387 & 0.958127 & 0.956647  \\
\hline
\end{tabular}
\end{table}

\begin{table}[htbp]
\caption{CNN-1D over AWID dataset: Classification results of each class. Total data records correctly classified: 548,489. Total number of data records: 575,643. Classification Accuracy = 95.28\%}
\label{tbl:cnnawid_correctsamples}
\centering
\begin{tabular}{ccccc}
\hline
 &	Estimated	& Correct	& Total	& F1 score  \\ \hline
normal    &      5615   &        4971   &    8097  &     72.5058 \\
flooding    &              43  &             38 &        20079 &    0.00377696 \\
injection  &             13185  &       13185  &   16682    &  88.2914 \\
impersonation &     556800  &     530295    & 530785 &  97.5179 \\
\hline
\end{tabular}
\end{table}

\begin{table}[htbp]
\caption{CNN-1D over AWID dataset: Classification results of each class in percentage. 
 Accuracy = 0.9528, F1 = 0.9351, Precision score = 0.9504, recall score = 0.9528}
\label{tbl:cnnawid_accuracy}
\centering
\begin{tabular}{ccccc}
\hline
Name &	Accuracy &	F1	& Precision &	Recall \\ \hline
 normal &      0.993451 &    0.725058  &     0.885307 &     0.613931 \\
flooding  &     0.965176&    0.00377696&   0.883721     &  0.00189252 \\
injection &      0.993925  &   0.882914     &    1      &           0.790373 \\
impersonation & 0.953105 &     0.975179&       0.952398 &     0.999077 \\
\hline
\end{tabular}
\end{table}

\end{document}